\documentclass[lettersize,journal]{IEEEtran}
\usepackage{amsmath,amsfonts}
\usepackage{algorithmic}
\usepackage{algorithm}
\usepackage{array}
\usepackage[caption=false,font=normalsize,labelfont=sf,textfont=sf]{subfig}
\usepackage{textcomp}
\usepackage{stfloats}
\usepackage{url}
\usepackage{verbatim}
\usepackage{graphicx}
\usepackage{cite}
\usepackage{hyperref}
\usepackage{multirow}
\usepackage{booktabs}
\usepackage{tabularx}
\usepackage{array}

\begin{document}

\title{PointGauss: Point Cloud-Guided Multi-Object Segmentation for Gaussian Splatting}

\author{Wentao Sun, Yiping Chen, John S. Zelek, Jonathan Li,~\IEEEmembership{Fellow,~IEEE}%
\thanks{W. Sun, J. S. Zelek, and J. Li are with the University of Waterloo, Waterloo, ON N2L 3G1, Canada (e-mails: w27sun@uwaterloo.ca; jzelek@uwaterloo.ca; junli@uwaterloo.ca).}%
\thanks{Y. Chen is with Sun Yat-Sen University, Zhuhai 519082, China (e-mail: chenyp79@mail.sysu.edu.cn).}%
}

\markboth{Journal of \LaTeX\ Class Files,~Vol.~*, No.~*, June~2026}%
{Shell \MakeLowercase{\textit{et al.}}: A Sample Article Using IEEEtran.cls for IEEE Journals}

\IEEEpubid{0000--0000/00\$00.00~\copyright~2026 IEEE}

\maketitle


\begin{abstract}
While 3D Gaussian Splatting (3DGS) has established new standards for high-fidelity 3D scene modeling, interpreting massive, unstructured Gaussian primitives into meaningful geospatial entities remains a critical challenge for remote sensing and urban applications. Existing 2D-to-3D distillation methods suffer from projection ambiguities, geometric inconsistencies, and computational bottlenecks when applied to complex, large-scale topographies. Furthermore, current benchmarks lack natively aligned 2D-3D ground truth tailored for splatting-based representations. To address these limitations, we propose \textit{PointGauss}, a unified 3D-native framework for robust semantic parsing and instance segmentation. By treating Gaussian primitives as unstructured point sets, PointGauss leverages Point Transformer V3 (PTv3) to extract scale-invariant geometric features directly from Gaussian attributes, effectively resolving projection ambiguities. To ensure scalability in building-scale scenes, we introduce an adaptive region-of-interest cropping strategy and an instance-aware distance-constrained rasterization pipeline for pixel-level, view-consistent projection. Additionally, we present \textbf{SplatSeg-360}, the first rigorous cross-scale benchmark tailored for 3DGS, comprising 32 complex scenes with full 360$^\circ$ coverage and over 6,300 natively aligned 2D-3D masks. Extensive experiments demonstrate that PointGauss operates in real-time and achieves state-of-the-art performance. Notably, it attains approximately 90\% 3D-mIoU in large-scale building scenarios and roughly 80\% 2D-mIoU in view-consistent 2D instance segmentation, outperforming baseline methods by 16\%. (\href{https://github.com/hbycswt/pointgauss}{Code}) 
\end{abstract}


\section{Introduction}
\label{sec:intro}
\IEEEPARstart{R}{ecent} advancements in 3D Gaussian Splatting (3DGS)~\cite{3dgs} have set new standards for high-fidelity 3D scene modeling. By representing scenes as a collection of explicit Gaussian primitives, 3DGS provides a direct link between visual synthesis and geometric spatial analysis. However, for remote sensing and urban applications, creating a visually realistic model is only the first step. The true challenge is semantic scene interpretation, which requires not only accurate reconstruction but also a deeper understanding of the environment. For tasks such as urban planning and digital twins, it is essential to decompose unstructured Gaussians into meaningful geospatial entities, such as buildings, vegetation, and vehicles. This decomposition provides semantic object representations for subsequent spatial analysis.

In this context, while NeRF-based segmentation~\cite{nerf-seg2,sa3d} has laid the groundwork, the research focus has rapidly shifted toward 3DGS-based frameworks~\cite{gaussian_grouping,SAGAgaussian,ClickGaussians} due to their explicit nature. These methods primarily follow a distillation-based paradigm, transferring semantic priors from 2D foundation models (e.g., SAM~\cite{sam}) onto Gaussian attributes. Despite their progress, they suffer from two critical technical bottlenecks when applied to complex geospatial environments:

(1) Geometric-Semantic Inconsistency in Complex Topography: By relying on lifting 2D features into 3D space, existing frameworks are inherently susceptible to projection ambiguities, which are exacerbated in urban scenes. In particular, when processing oblique imagery, 2D masks often fail to align precisely with the vertical structures and bases of buildings. This leads to severe semantic leakage across boundaries and the misclassification of off-surface floaters—low-density Gaussians that inherit incorrect semantic scores from background pixels. Without explicit 3D structural constraints, maintaining a consistent segmentation that respects the underlying geometry of large-scale infrastructures remains challenging.

(2) Scalability Bottlenecks and Computational Overhead: The reliance on multi-stage distillation and cross-view feature alignment necessitates exhaustive iterative optimization. While manageable for tabletop objects, this paradigm becomes computationally prohibitive for large-scale building complexes where the number of Gaussian primitives increases exponentially. The resulting memory footprint and training time hinder the rapid deployment required for time-sensitive remote sensing tasks. Furthermore, these methods often lack a unified architecture capable of performing direct, high-fidelity geometric reasoning on Gaussians, failing to maintain structural coherence across massive point sets while satisfying both autonomous parsing and interactive extraction.

\IEEEpubidadjcol

Additionally, existing benchmarks for 3D segmentation in Gaussian Splatting, such as NVOS~\cite{nvos}, Spin-NeRF~\cite{spinerf}, and LERF-Mask~\cite{gaussian_grouping}, exhibit several critical limitations that hinder rigorous evaluation: (1) Limited Scene Complexity: Most focus on single-object scenarios, failing to represent the multi-object occlusions typical of real-world environments. (2) Incomplete View Coverage: Constrained camera trajectories with partial angular coverage prevent the assessment of 3DGS's unique capability for globally consistent 360$^\circ$ reconstruction. (3) Representation Incompatibility: Crucially, standard 3D datasets like ScanNet~\cite{scannet} or S3DIS~\cite{s3dis} are ill-suited for 3DGS. While these datasets provide mesh or point cloud labels derived from depth sensors, 3DGS consists of millions of anisotropic Gaussian primitives with rich attributes (e.g., opacity, spherical harmonics). These primitives do not align one-to-one with rigid 3D geometries, making existing labels geometrically inconsistent for splatting-based representations. (4) Lack of Unified 2D-3D Ground Truth: In many existing benchmarks, 2D masks are merely coarse projections of 3D labels, which often lack the pixel-level precision required to evaluate high-fidelity 3DGS renderings. A benchmark that provides natively aligned, high-resolution 2D and 3D instance annotations for Gaussian-based representations is still absent.

To address these challenges, we propose PointGauss, a unified, geometry-centric framework designed for robust semantic parsing and instance segmentation across diverse spatial scales. Our core innovation lies in treating Gaussian primitives as an unstructured point set and leveraging the Point Transformer V3 (PTv3) \cite{PTV3} backbone to perform high-fidelity 3D feature extraction directly from Gaussian centroids and attributes. Unlike 2D-based methods that depend on perspective-specific priors, this 3D-native approach allows our model to capture scale-invariant geometric patterns, ranging from centimeter-level indoor objects to decameter-level urban infrastructures. To handle the increased point density and spatial complexity of building-scale scenes, we implement a region-of-interest (ROI) cropping strategy and adaptive point batching. These techniques enable the rapid, single-pass initialization of 3D semantic fields while maintaining the structural integrity of large-scale geospatial entities. Furthermore, we introduce an instance-aware splatting projection pipeline that utilizes distance-constrained rasterization to efficiently propagate labels to novel viewpoints with pixel-level precision. This approach ensures consistent segmentation performance across both constrained indoor and expansive outdoor environments.

To facilitate rigorous evaluation, we present \textbf{SplatSeg-360}, a comprehensive benchmark featuring complex multi-object scenes with full 360$^\circ$ coverage and over 6,300 natively aligned 2D-3D masks. Encompassing 32 diverse scenes across four specialized sub-datasets (\textit{Residence360}, \textit{Campus360}, \textit{Vehicles360}, and \textit{DesktopObjects360}), the benchmark provides a rigorous cross-scale spectrum—ranging from macro-scale urban architectures to micro-scale indoor clutter. Extensive experiments demonstrate that our proposed algorithm effectively achieves highly accurate 3D segmentation and view-consistent 2D segmentation across all these urban, street, and indoor desktop environments while operating in real-time. Notably, in large-scale building scenarios, our method achieves approximately 90\% 3D-mIoU. Furthermore, in view-consistent 2D instance segmentation tasks, it attains a 2D-mIoU of roughly 80\%, outperforming the baseline methods by 16\%.

In summary, the main contributions of this article are as follows:
\begin{itemize}
    \item \textbf{A Unified 3D-Native Segmentation Framework}: We propose \textit{PointGauss}, which extracts scale-invariant features directly from unstructured Gaussian attributes using Point Transformer V3 (PTv3). This geometry-centric approach inherently resolves the projection ambiguities and semantic inconsistencies of traditional 2D-to-3D distillation paradigms.
    
    \item \textbf{Scalable and Instance-Aware Rendering Pipeline}: To overcome computational bottlenecks in large-scale urban scenes, we introduce an adaptive ROI cropping strategy for rapid 3D semantic initialization. Additionally, our distance-constrained rasterization ensures strict pixel-level precision and view consistency during label projection.
    
    \item \textbf{A Comprehensive Cross-Scale Benchmark}: We construct \textbf{SplatSeg-360}, the first rigorous dataset explicitly tailored for Gaussian representations. It provides 32 diverse scenes with full 360$^{\circ}$ coverage and over 6,300 natively aligned 2D-3D masks, bridging the evaluation gap from micro-scale indoor clutter to macro-scale infrastructures.
    
    \item \textbf{State-of-the-Art Performance}: Operating in real-time, PointGauss achieves superior structural coherence, yielding an exceptional 90\% 3D-mIoU in large-scale building scenarios. For view-consistent 2D instance segmentation, it attains approximately 80\% 2D-mIoU, outperforming baseline methods by 16\%.
\end{itemize}


\section{Related Work}
\label{sec:related-work}
\subsection{3D Gaussian Representations}
Recent progress in 3D Gaussian Splatting (3DGS) has led to diverse methodological improvements, which we categorize into seven key categories following \cite{3dgs-survey}: (i) sparse view reconstruction~\cite{3dgs4Img,sprs_DNGaussian,sprs_fsgs,sprs_pxlsplt}, (ii) memory-efficient representations~\cite{mem_cmprs3dgs,colorcompact,mem_gsimg}, (iii) photorealistic rendering techniques~\cite{phtrel-GaussianShader,phtrel-multiscale,phtrel-gsc,phtrel-mipsplt}, (iv) optimization strategies~\cite{Sugar,imp-FreGS,2dgs,imp-Scaffold}, (v) attribute learning~\cite{gaussian_grouping,property-lang,feature3dgs}, (vi) hybrid architectures~\cite{hybrid-4dgs,hybrid-Deformable}, and (vii) rendering innovations~\cite{rendering-RayT,rendering-tracing}. 

Among these, optimization remains a particularly active area, with recent works addressing the persistent issue of imbalanced reconstruction quality across different regions of a scene. Geometry-aware approaches~\cite{2dgs,imp-Scaffold} have demonstrated notable success by incorporating structural priors into the optimization process. For example, 2DGS\cite{2dgs} introduces planar Gaussian primitives to better capture surface continuity while reducing computational overhead.

\subsection{Segmentation for Gaussian Splatting}
Several notable works have advanced this field, such as Feature3DGS~\cite{feature3dgs}, Gaussian Grouping~\cite{gaussian_grouping}, OmniSeg3D~\cite{omniseg}, SAGA~\cite{SAGAgaussian}, and Click-Gaussian~\cite{ClickGaussians}. These methods generally follow a similar pipeline: they extract 2D masks or features using SAM, lift the information into 3D space via contrastive learning or distillation, and then project the resulting 3D segmentation features back to 2D to enable segmentation from novel viewpoints. However, each method varies in how it processes SAM-derived 2D information and faces distinct limitations.

OmniSeg3D addresses 2D masks' view inconsistency through hierarchical contrastive learning, yet struggles with scale adaptability. SAGA introduces scale-gated affinity features for multi-level cues but overlooks 2D mask conflicts. Click-Gaussian improves multi-view consistency with feature smoothing but is limited in handling multi-instance Gaussian primitives effectively.

Beyond individual shortcomings, these approaches share several common challenges: underutilization of the inherent 3D spatial structure, increased architectural complexity due to multi-stage pipelines, and a strong dependency on 2D segmentation models.

These limitations motivate our proposed approach: a point cloud-guided 3D segmentation framework that leverages the native geometry of Gaussian primitives, eliminates reliance on 2D masks, and improves both efficiency and multi-view consistency.

\subsection{Point Cloud Segmentation}
Point cloud segmentation encompasses both semantic segmentation and instance segmentation. Semantic segmentation methods can be categorized based on their processing strategies. Point-based approaches, such as PointNet++~\cite{pointnet++}, operate directly on raw 3D coordinates. Voxel-based methods~\cite{voxelPts} discretize point clouds into volumetric grids for structured processing, while multi-view techniques~\cite{m-viewPts} project 3D data onto 2D image planes to leverage mature 2D CNN architectures. Hybrid architectures, such as PointTransformerV3 (PTV3)~\cite{PTV3}, combine these strategies to improve scalability and efficiency for large-scale scenes.

Instance segmentation builds upon semantic segmentation by distinguishing individual object instances. Proposal-based methods, such as 3D-SIS~\cite{3d-sis}, generate region proposals to localize instances, while CRSNet~\cite{CRSNet} incorporate user-provided cues. Clustering-based methods, like SGPN~\cite{sgpn}, learn similarity matrices to group points into distinct instances. 

Our approach adopts a point-based segmentation strategy while leveraging the unique properties of 3D Gaussian Splatting (3DGS) to enhance spatial coherence and enable real-time multi-object segmentation.

\begin{figure*}[t!]
    \centering
    \includegraphics[width=1.0\linewidth]{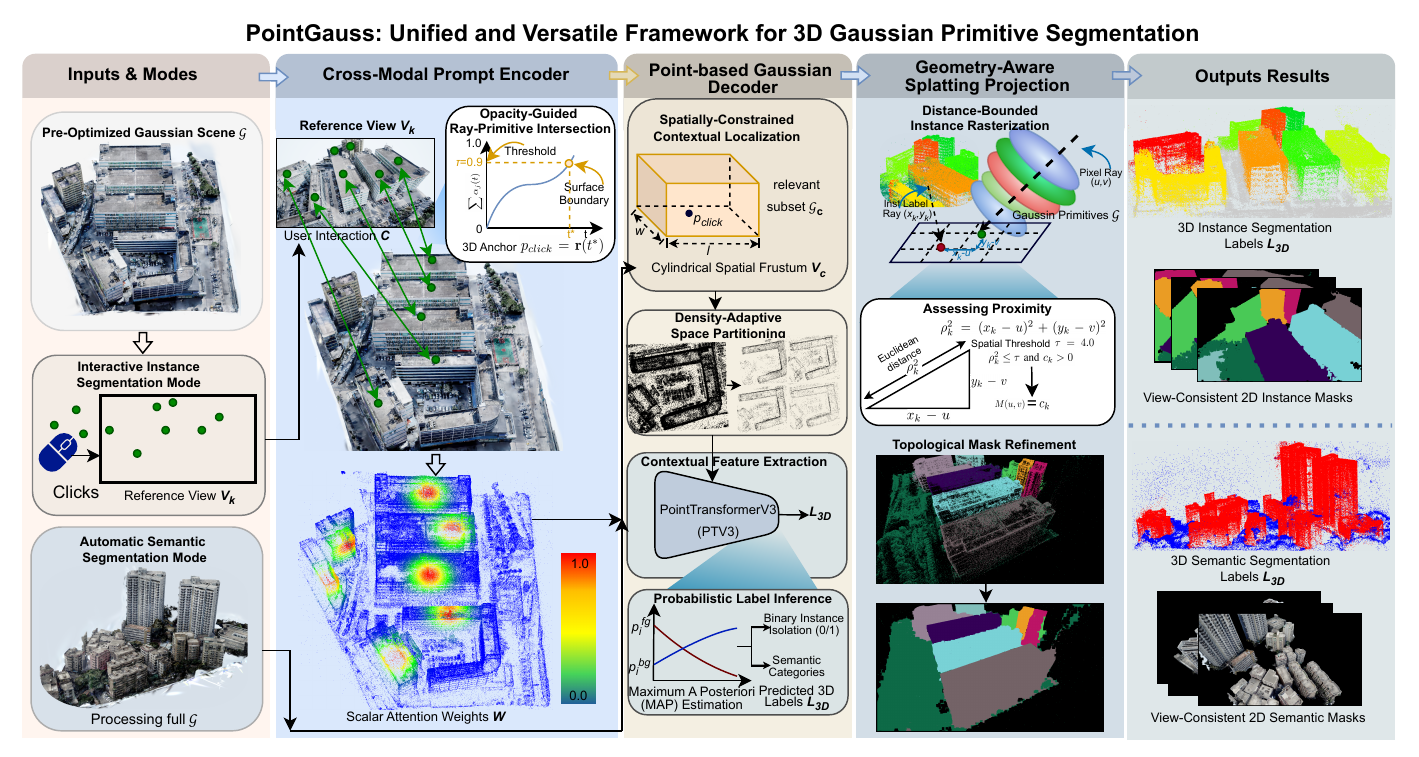}
    \caption{\textbf{Overall architecture of the proposed PointGauss framework.} Given a pre-optimized 3D Gaussian scene $\mathcal{G}$ and optional sparse user interactions $\mathcal{C}$, the framework operates in either automatic semantic or interactive instance segmentation modes. It consists of three core modules: (1) A \textbf{Cross-Modal Prompt Encoder} maps 2D clicks into 3D spatial attention weights via an opacity-guided ray-primitive intersection and distance-decaying attention mechanism. (2) A \textbf{Point-based Gaussian Decoder} processes feature-augmented Gaussians $\mathcal{G}^*$ through spatially-constrained localization and density-adaptive partitioning, utilizing PointTransformerV3 to infer primitive-level 3D labels $\mathcal{L}_{3D}$. (3) A \textbf{Geometry-Aware Splatting Projection} module employs distance-bounded instance rasterization and topological refinement to render view-consistent 2D masks $\mathcal{M}_{2D}$ from novel viewpoints.}
    \label{fig:architecture}
\end{figure*}

\section{Proposed Method} 
\label{sec:method} 
As illustrated in Fig.~\ref{fig:architecture}, PointGauss is conceptualized as a unified and versatile framework for the segmentation of 3D Gaussian primitives. The architecture is meticulously designed around three core modules: a \textit{Cross-Modal Prompt Encoder} that bridges 2D visual interactions with 3D geometric spaces, a \textit{Point-based Gaussian Decoder} for high-fidelity 3D label prediction, and a \textit{Geometry-Aware Splatting Projection} module for generating view-consistent 2D masks. To accommodate diverse analytical requirements, our framework operates under two distinct paradigms: an \textbf{automatic semantic segmentation} mode for holistic scene understanding, and an \textbf{interactive instance segmentation} mode, in which user-guided prompts spatially constrain and direct the segmentation process.

\subsection{Problem Formulation}
\label{subsec:problem-definition}
Given a pre-optimized 3D Gaussian scene $\mathcal{G}=\{g_i\}_{i=1}^N$ (reconstructed via 2DGS~\cite{2dgs}), each Gaussian primitive $g_i=(\mu_i,\Sigma_i,\alpha_i,c_i)$ is fully parameterized by its spatial center $\mu_i\in\mathbb{R}^3$, anisotropic covariance $\Sigma_i\in\mathbb{S}^3_{++}$, opacity $\alpha_i\in[0,1]$, and appearance feature $c_i$. For the interactive paradigm, a set of sparse user interactions $\mathcal{C}=\{c_j\}_{j=1}^M$ (e.g., point clicks) in a reference view $V_k$ is introduced as an auxiliary spatial prior. 

Our primary objective is to formulate a mapping function $\mathcal{F}$ that simultaneously derives primitive-level 3D annotations and view-consistent 2D projections:
\begin{equation}
    \mathcal{F}:(\mathcal{G},\mathcal{C})\rightarrow(\mathcal{L}_{3D},\mathcal{M}_{2D})
    \label{eq:mapping}
\end{equation}
where $\mathcal{L}_{3D}=\{l_i\}_{i=1}^N$ denotes the predicted category or instance labels for the Gaussian scene. Specifically, $l_i\in\{1,\dots,K\}$ for semantic parsing ($\mathcal{C}=\emptyset$), and $l_i\in\{0,1\}$ for binary instance isolation ($\mathcal{C}\neq\emptyset$). The term $\mathcal{M}_{2D}=\Psi(\mathcal{L}_{3D},V)$ represents the 2D rendered mask via our differentiable projection operator $\Psi(\cdot)$. A robust $\mathcal{F}$ must satisfy two fundamental constraints: \textbf{Prompt Faithfulness}, ensuring $\mathcal{L}_{3D}$ strictly aligns with the intent of $\mathcal{C}$ in the reference view, and \textbf{3D Geometric Coherence}, guaranteeing that the resulting masks remain spatially and temporally consistent across arbitrary novel viewpoints.

\subsection{Cross-Modal Prompt Encoder}
The prompt encoder is explicitly designed to translate 2D sparse user clicks into dense 3D spatial attention weights. When an interaction occurs at pixel $\mathbf{p}=(u,v)$, we trace a viewing ray $\mathbf{r}(t)=\mathbf{o}+t\mathbf{d}$ originating from the camera center $\mathbf{o}$ along the direction $\mathbf{d}$.

To precisely localize the interaction within the 3D space, we formulate an opacity-guided ray-primitive intersection mechanism. The optimal intersection depth $t^*$ is determined by accumulating the opacity along the ray until a rigid surface boundary is reached:
\begin{equation}
    t^*=\underset{t}{\arg\min}\left\{t\;\bigg|\;\sum_j\alpha_j(t)>\tau\right\}
    \label{eq:intersection}
\end{equation}
where $\alpha_j(t)$ is the localized opacity contribution at depth $t$, and the threshold $\tau=0.9$ ensures reliable surface penetration. The derived 3D anchor $p_{click}=\mathbf{r}(t^*)$ acts as the origin for spatial influence.

Subsequently, we model the spatial relevance of all surrounding Gaussians as a distance-decaying attention mechanism. For any primitive $G_i$ with mean $\boldsymbol{\mu}_i$, its relevance weight $w_i$ is computed as:
\begin{equation}
    w_i=\exp\left(-\frac{\|\boldsymbol{\mu}_i-p_{click}\|^2}{2\sigma^2}\right)
    \label{eq:calweight}
\end{equation}
where the bandwidth parameter $\sigma$ (empirically set to $0.15\,\text{m}$) dictates the decay rate. This operation yields a smooth, continuous scalar field over the Gaussian scene, inherently filtering out geometrically irrelevant primitives. The calculated weight vector $W=\{w_i\}_{i=1}^N$ is explicitly concatenated to the geometric attributes of the Gaussians, forming a feature-augmented representation $\mathcal{G}^*$.

\subsection{Gaussian Decoder}
Leveraging the augmented primitives $\mathcal{G}^*$, the Gaussian Decoder executes a coarse-to-fine segmentation pipeline, deeply integrating spatial priors and local geometries.

\textbf{Spatially-Constrained Contextual Localization.} To eliminate computational redundancy and focus the network on the target instance, we define an unconstrained-height cuboid volume $V_c$ anchored at $p_{click}$. The user-defined length $l$ and width $w$ are adaptively scaled to exceed the dimensions of the target object. The subset of relevant Gaussians $\mathcal{G}_c$ is geometrically filtered by:
\begin{equation}
    \mathcal{G}_c=\left\{g_i^*\in\mathcal{G}^* \;\mid\; |x_i-p_{click}^x|\leq\frac{l}{2} \land |y_i-p_{click}^y|\leq\frac{w}{2}\right\}
    \label{eq:selGaussian}
\end{equation}
where the user-specified bounds $l$ and $w$ define the horizontal spatial receptive field, capturing the full vertical extent and topological integrity of the targeted instance within the expanded regional column.

\textbf{Density-Adaptive Space Partitioning.} Given the highly variable point density inherent to Gaussian splatting, standard point-based networks suffer from memory bottlenecks. For the localized point set $\mathcal{P}_c$ derived from $\mathcal{G}_c$, we employ a density-adaptive partitioning strategy:
\begin{equation}
    \mathcal{P}_c=\bigcup_{k=1}^{\lceil|\mathcal{P}_c|/N_{max}\rceil}\mathcal{B}_k
    \label{eq:ptsbatch}
\end{equation}
where the point cloud is decomposed into mutually exclusive batches $\mathcal{B}_k$, strictly bounding the local density to $|\mathcal{B}_k|\leq N_{max}$ ($N_{max}=8192$). This ensures stable gradient propagation and consistent topological feature extraction. 

\textbf{Contextual Feature Extraction.} The partitioned data is ingested by PointTransformerV3 (PTV3)~\cite{PTV3}, which excels at capturing complex local-global geometric correlations. PTV3 functions as our primary feature extractor, outputting high-dimensional point-wise semantic logits.

\textbf{Probabilistic Label Inference.} The final instance discretization is achieved through Maximum A Posteriori (MAP) estimation over the predicted logits. The binary label $inst\_label_i$ is assigned as:
\begin{equation}
    inst\_label_i=\mathbb{I}\left(p_i^{fg}>p_i^{bg}\right)
    \label{eq:instlabel}
\end{equation}
where $p_i^{fg}$ and $p_i^{bg}$ represent the categorical probabilities, and $\mathbb{I}(\cdot)$ denotes the indicator function.

\subsection{Geometry-Aware Splatting Projection}
\label{sec:splatting}
To translate the 3D discrete labels $\mathcal{L}_{3D}$ back to arbitrary 2D observation planes, we introduce an instance-aware rasterization module. 

\textbf{Distance-Bounded Instance Rasterization.} Unlike traditional color splatting, instance labels require hard assignment to preserve sharp boundaries. For a target pixel $(u,v)$, the projected label $M(u,v)$ is determined by assessing the spatial proximity of projected Gaussians in the image plane:
\begin{equation}
M(u,v)=\begin{cases}c_k&\text{if }\exists k\in\mathcal{N}(u,v),\ \rho^2_k\leq\tau\text{ and }c_k>0\\0&\text{otherwise}\end{cases}
\label{eq:rendering}
\end{equation}
where $\mathcal{N}(u,v)$ represents the set of Gaussians intersecting the pixel ray. The term $\rho^2_k=(x_k-u)^2+(y_k-v)^2$ quantifies the 2D Euclidean divergence, where $(x_k, y_k)$ denotes the projected coordinate of the $k$-th Gaussian primitive's center on the image plane, and $\tau=4.0$ acts as a strict spatial threshold. By terminating the rasterization pass at the first valid instance label that satisfies the proximity constraint, we achieve computationally efficient label propagation while rigorously maintaining structural boundaries.

\begin{algorithm}[H]
\caption{Instance Label Projection in Rasterization}
\label{alg:instance_projection}
\begin{algorithmic}
\STATE 
\STATE {\textsc{ProjectInstanceLabels}}$(\;pix,$ 
\STATE \hspace{2.0cm}$collected\_id,$ 
\STATE \hspace{2.0cm}$inst\_label,$ 
\STATE \hspace{2.0cm}$inst\_image\;)$
\STATE \hspace{0.5cm}$ \rho2d\_threshold \gets 4.0 $
\STATE \hspace{0.5cm}\textbf{for each} pixel $(x,y)$ \textbf{in} tile
\STATE \hspace{1.0cm}$ inst\_image[x,y] \gets 0 $
\STATE \hspace{1.0cm}\textbf{for each} Gaussian $j$ \textbf{in} primitives
\STATE \hspace{1.5cm}$ prim\_id \gets collected\_id[j] $
\STATE \hspace{1.5cm}$ (dx,dy) \gets points\_xy\_image[prim\_id] - (x,y) $
\STATE \hspace{1.5cm}$ \rho2d \gets dx^2 + dy^2 $
\STATE \hspace{1.5cm}\textbf{if } $ \rho2d \leq \rho2d\_threshold $
\STATE \hspace{2.0cm}$ current \gets inst\_label[prim\_id] $
\STATE \hspace{2.0cm}\textbf{if } $ current > 0 \wedge inst\_image[x,y]=0 $
\STATE \hspace{2.5cm}$ inst\_image[x,y] \gets current $
\STATE \hspace{2.0cm}\textbf{end if}
\STATE \hspace{1.5cm}\textbf{end if}
\STATE \hspace{1.0cm}\textbf{end for}
\STATE \hspace{0.5cm}\textbf{end for}
\STATE \hspace{0.5cm}\textbf{return} $inst\_image$
\end{algorithmic}
\end{algorithm}

\textbf{Topological Mask Refinement.} Due to the discrete nature of 3D-to-2D projection, minor morphological artifacts (e.g., localized cavities or jagged contours) may arise. We mitigate these via a cascaded topological refinement: (1) multi-scale morphological closing to heal boundary discontinuities, (2) edge-aware cavity filling to preserve internal structural coherence, and (3) primary connected-component extraction to eliminate isolated spatial noise. 

\subsection{Summary}
PointGauss establishes a new paradigm for segmentation in Gaussian Splatting scenes by directly operating on Gaussian primitives, ensuring native compatibility with Gaussian Splatting rendering and preserving geometric fidelity. The framework comprises three core components: (1) a cross-modal prompt encoder that bridges 2D user interactions and 3D Gaussian representations, (2) a Gaussian decoder for generating segmentation outputs, and (3) a rendering module that ensures view-consistent results. This architecture achieves real-time performance while enabling accurate, geometry-aware interactive segmentation.

\begin{figure*}[t]
    \centering
    \includegraphics[width=\textwidth]{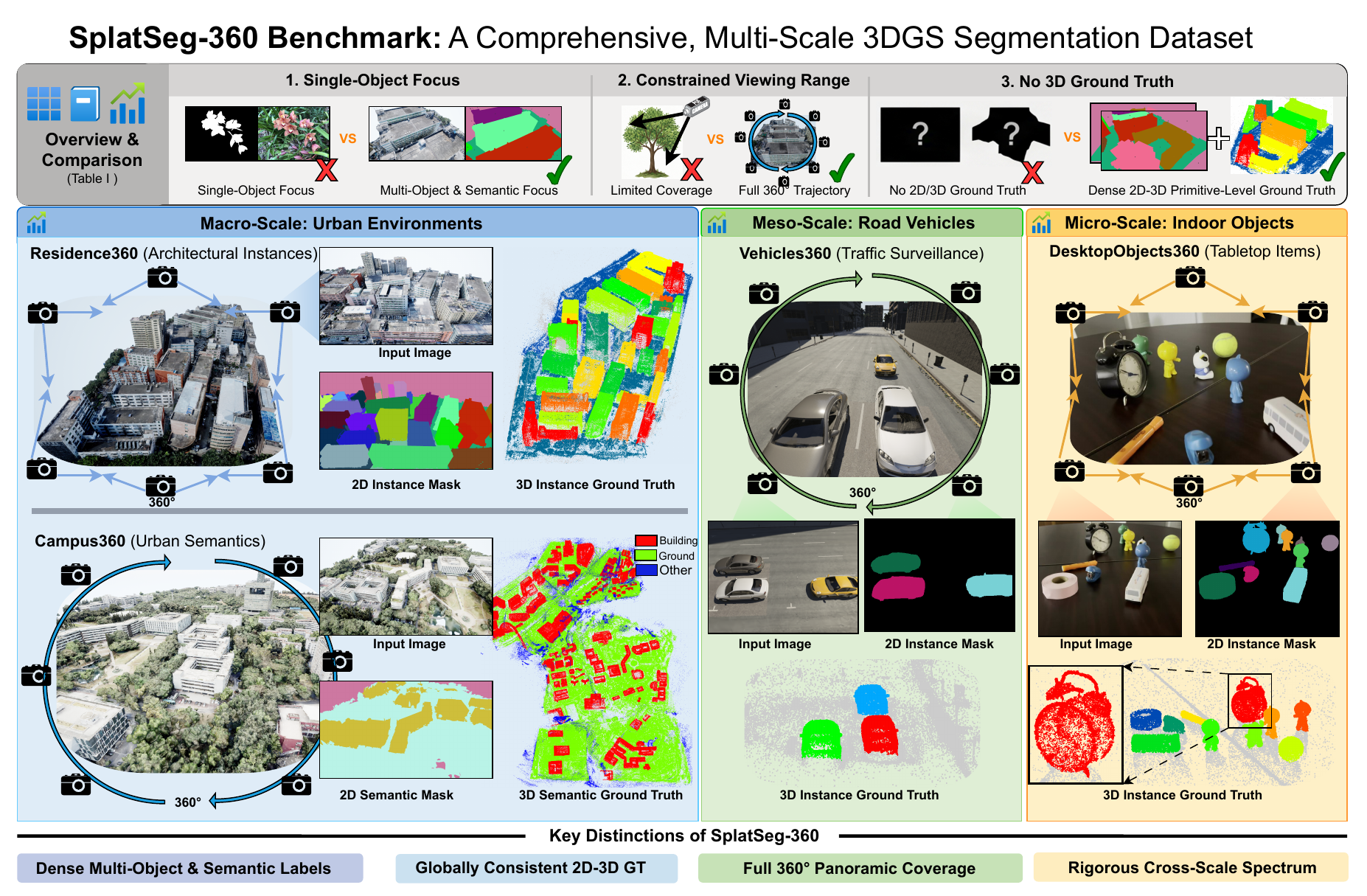} 
    \caption{\textbf{Overview of the proposed SplatSeg-360 benchmark.} The top panel illustrates our motivation by contrasting the inherent limitations of existing datasets (i.e., single-object focus, constrained viewing ranges, and absence of 3D ground truth) against our comprehensive approach. The main panels showcase the cross-scale paradigm of SplatSeg-360 across four diverse sub-datasets: Residence360 and Campus360 (macro-scale urban environments), Vehicles360 (meso-scale road elements), and DesktopObjects360 (micro-scale indoor clutter). For each scenario, the benchmark uniquely provides full 360$^\circ$ camera trajectories alongside globally consistent 2D instance/semantic masks and dense 3D primitive-level annotations.}
    \label{fig:dataset_overview}
\end{figure*}

\section{The SplatSeg-360 Benchmark}
\label{sec:dataset}

\subsection{Motivation and Limitations of Existing Benchmarks}
High-quality, multi-view semantic and instance annotations are indispensable for training and evaluating 3D Gaussian Splatting (3DGS) segmentation algorithms. However, existing standard datasets, such as NVOS~\cite{nvos}, Spin-NeRF~\cite{spinerf}, and LERF-Mask~\cite{gaussian_grouping}, suffer from several inherent limitations that constrain their applicability to comprehensive 3D spatial understanding:
\begin{enumerate}
    \item \textbf{Single-Object Focus:} Most frameworks only annotate a single, isolated foreground object, failing to replicate complex multi-target environments.
    \item \textbf{Constrained Viewing Range:} Many benchmarks are restricted to forward-facing or bounded multi-view captures, omitting complete 360$^\circ$ bounded or unbounded trajectories.
    \item \textbf{Absence of 3D Ground Truth:} Existing benchmarks predominantly provide 2D image-space masks, missing dense, primitive-level 3D annotations. This is highly problematic for advanced point-cloud or primitive-based frameworks (like PointGauss), which rely heavily on direct 3D structural supervision.
\end{enumerate}

To bridge these critical gaps, we introduce \textbf{SplatSeg-360} (\textbf{Fig.~\ref{fig:dataset_overview}}), a comprehensive, multi-scale, full-coverage 3DGS segmentation benchmark. As summarized in Table~\ref{tab:dataset_comparison}, SplatSeg-360 distinguishes itself by providing: (i) dense multi-object and semantic annotations, (ii) globally consistent 2D-3D cross-view ground truth, (iii) full 360$^\circ$ panoramic coverage, and (iv) a rigorous cross-scale spectrum encompassing macro-scale urban architectures, meso-scale road traffic, and micro-scale tabletop items.

\begin{table*}[t]
\caption{Comparative Analysis of SplatSeg-360 Against Existing 3D Scene Segmentation Benchmarks.}
\label{tab:dataset_comparison}
\centering
\small
\resizebox{\textwidth}{!}{%
\setlength{\tabcolsep}{4.5pt}
\begin{tabular}{lccccccccc}
\toprule
\textbf{Dataset} & \textbf{\# Scenes} & \textbf{Images / Scene} & \textbf{Target Type} & \textbf{360$^\circ$ Coverage} & \textbf{2D Ann.} & \textbf{3D Ann.} & \textbf{Calibration} & \textbf{Environment} & \textbf{Spatial Scale} \\
\midrule
NVOS~\cite{nvos} & 7 & $\sim$40 & Single-Obj. & No & Yes & No & Yes & Real (Outdoor) & Small \\
Spin-NeRF~\cite{spinerf} & 10 & $\sim$60 & Single-Obj. & No & Yes & No & Yes & Real (Bi-domain) & Small \\
LERF-Mask~\cite{gaussian_grouping} & 3 & $\sim$200 & Multi-Obj. & Yes & Yes & No & Yes & Real (Indoor) & Small \\
ScanNet~\cite{scannet} & $>$1500 & $>$1000 & Semantic & Yes & No & Yes & Yes & Real (Indoor) & Small \\
Mip-NeRF360~\cite{mipnerf360} & 9 & $\sim$300 & None & Yes & No & No & Yes & Real (Bi-domain) & Small \\
NeRDs360~\cite{nerds360} & 71 & $\sim$200 & Multi-Obj. & Yes & Yes & No & Yes & Synthetic (Outdoor) & Medium \\
\midrule
\textbf{SplatSeg-360 (Ours)} & \textbf{32} & \textbf{36--957} & \textbf{Multi-Obj. \& Sem.} & \textbf{Yes} & \textbf{Yes} & \textbf{Yes} & \textbf{Yes} & \textbf{Real + Synth.} & \textbf{Small, Medium, Large} \\
\bottomrule
\end{tabular}
}
\end{table*}


\begin{table}[h]
\caption{Detailed Statistical Layout of the SplatSeg-360 Benchmark.}
\label{tab:sub_dataset_stats}
\centering
\footnotesize
\setlength{\tabcolsep}{3.5pt} 
\begin{tabularx}{\columnwidth}{l c >{\raggedright\arraybackslash}X c c}
\toprule
\textbf{Sub-dataset} & \textbf{ID} & \textbf{Scene Name} & \textbf{\# Images} & \textbf{\# Instances} \\
\midrule
\multirow{5}{*}{Residance360} 
 & 1 & Scene 1 & 81 & 9 \\
 & 2 & Scene 2 & 66 & 10 \\
 & 3 & Scene 3 & 103 & 7 \\
 & 4 & Scene 4 & 68 & 7 \\
 & 5 & Scene 5 & 71 & 7 \\
\midrule
\multirow{11}{*}{Campus360} 
 & 1 & Campus 1 & 54 & -- (Semantic) \\
 & 2 & Campus 2 & 61 & -- (Semantic) \\
 & 3 & Campus 3 & 35 & -- (Semantic) \\
 & 4 & Campus 4 & 70 & -- (Semantic) \\
 & 5 & Campus 5 & 88 & -- (Semantic) \\
 & 6 & Campus 6 & 50 & -- (Semantic) \\
 & 7 & Campus 7 & 142 & -- (Semantic) \\
 & 8 & Campus 8 & 46 & -- (Semantic) \\
 & 9 & Campus 9 & 41 & -- (Semantic) \\
 & 10 & Campus 10 & 36 & -- (Semantic) \\
 & 11 & Campus 11 & 45 & -- (Semantic) \\
\midrule
\multirow{10}{*}{Vehicles360} 
 & 1 & SF\_6thAnd\allowbreak Mission\_med0 & 198 & 4 \\
 & 2 & SF\_6thAnd\allowbreak Mission\_med6 & 192 & 3 \\
 & 3 & SF\_6thAnd\allowbreak Mission\_med7 & 199 & 3 \\
 & 4 & SF\_6thAnd\allowbreak Mission\_med10 & 199 & 3 \\
 & 5 & SF\_6thAnd\allowbreak Mission\_med12 & 198 & 4 \\
 & 6 & SF\_GrantAnd\allowbreak California1 & 199 & 4 \\
 & 7 & SF\_GrantAnd\allowbreak California2 & 195 & 4 \\
 & 8 & SF\_GrantAnd\allowbreak California3 & 196 & 3 \\
 & 9 & SF\_VanNessAnd\allowbreak Turk3 & 192 & 4 \\
 & 10 & SF\_VanNessAnd\allowbreak Turk5 & 199 & 3 \\
\midrule
\multirow{6}{*}{DesktopObjects360} 
 & 1 & Desk 1 & 324 & 10 \\
 & 2 & Desk 2 & 373 & 10 \\
 & 3 & Desk 3 & 957 & 10 \\
 & 4 & Desk 4 & 740 & 9 \\
 & 5 & Desk 5 & 764 & 10 \\
 & 6 & Desk 6 & 206 & 7 \\
\bottomrule
\end{tabularx}
\end{table}

\subsection{Sub-dataset Composition and Data Processing}
The SplatSeg-360 benchmark aggregates 32 diverse scenarios distributed across four specialized sub-datasets, intentionally curated to represent distinct geometric complexities and physical scales. The detailed structure is specified below:

\subsubsection{\textbf{Residence360 (Macro-Scale Architectural Instances)}}
Sourced from the photogrammetric reconstructions of the \textit{UrbanScene3D} dataset, this real-world outdoor sub-dataset consists of 5 large-scale residential zones. To render it compatible with 3DGS pipelines, we re-sampled the multi-view trajectories and generated rigorous COLMAP-compatible camera calibrations. More importantly, we performed dense, manual 3D bounding-primitive segmentation to isolate individual architectural structures, providing precise multi-object instance labels across 5 independent scenes.

\subsubsection{\textbf{Campus360 (Macro-Scale Urban Semantics)}}
Capturing expansive educational complexes from the \textit{ShenzhenUniversity} region of \textit{UrbanScene3D}, this sub-dataset contains 11 distinct real-world scenes. It is tailored specifically for fully automated 3D semantic mapping. We applied meticulous spatial cropping and manual 3D parsing to cleanly separate land-cover assets into two major semantic categories: \textit{Building} structures and static \textit{Background} (encompassing terrain, roads, and sky grids). Full COLMAP camera optimization is integrated natively.

\subsubsection{\textbf{Vehicles360 (Meso-Scale Road Elements)}}
To evaluate algorithm efficacy on mid-sized foreground targets within standard traffic surveillance contexts, we integrated 10 street-view 3D Gaussian scenes source-sampled from the \textit{NeRDs360} dataset. Leveraging these high-fidelity continuous trajectories, we manually re-annotated the 3D primitives with dense vehicle instance masks (ranging from 3 to 4 independent vehicles per scene), thereby bypassing 2D-only limitations and creating a robust baseline for 3D-space cross-view identity-consistent tracking and segmentation.

\subsubsection{\textbf{DesktopObjects360 (Micro-Scale Indoor Clutter)}}
To complete the cross-scale paradigm, this real-world sub-dataset introduces 6 highly cluttered, close-up tabletop environments. Each scene features numerous small objects characterized by complex geometry boundaries, mutual occlusion, and specular surface textures. To facilitate multi-dimensional evaluation, we establish dense, primitive-level 3D instance annotations for every target item, complemented by temporally-consistent 2D instance segmentation masks across all video frames. This dual-domain labeling creates a challenging environment for localized boundary delineation.


The exhaustive scene-by-scene statistics for all four sub-datasets are cataloged in Table~\ref{tab:sub_dataset_stats}.

\begin{figure*}
    \centering
    \includegraphics[width=\linewidth]{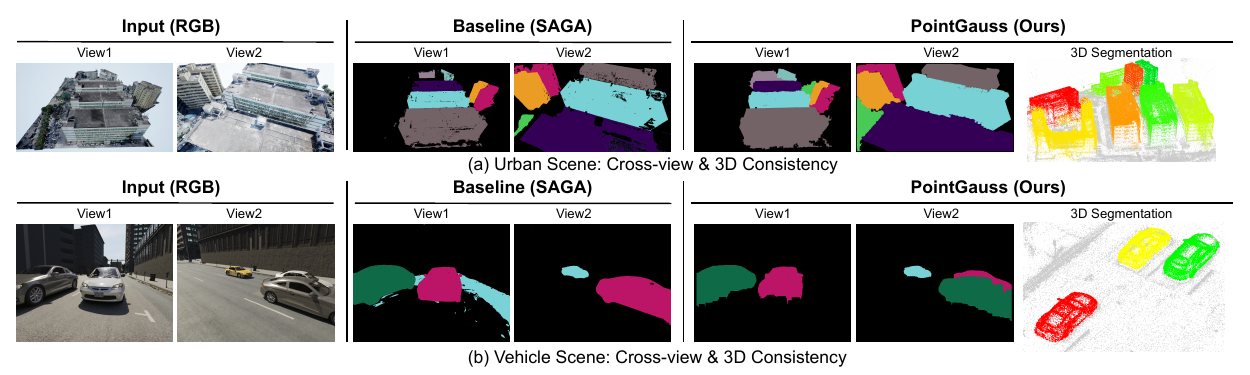}
    \caption{Qualitative comparison of 3D scene segmentation results on the Residence360 and Vehicles360 sub-datasets. Different colors denote different instances. The baseline method (SAGA) tends to yield incomplete segmentations for large-scale houses, whereas our method exhibits jagged boundaries on small-to-medium-sized vehicles.}
    \label{fig:urbanInst}
\end{figure*}

\begin{figure*}
    \centering
    \includegraphics[width=\linewidth]{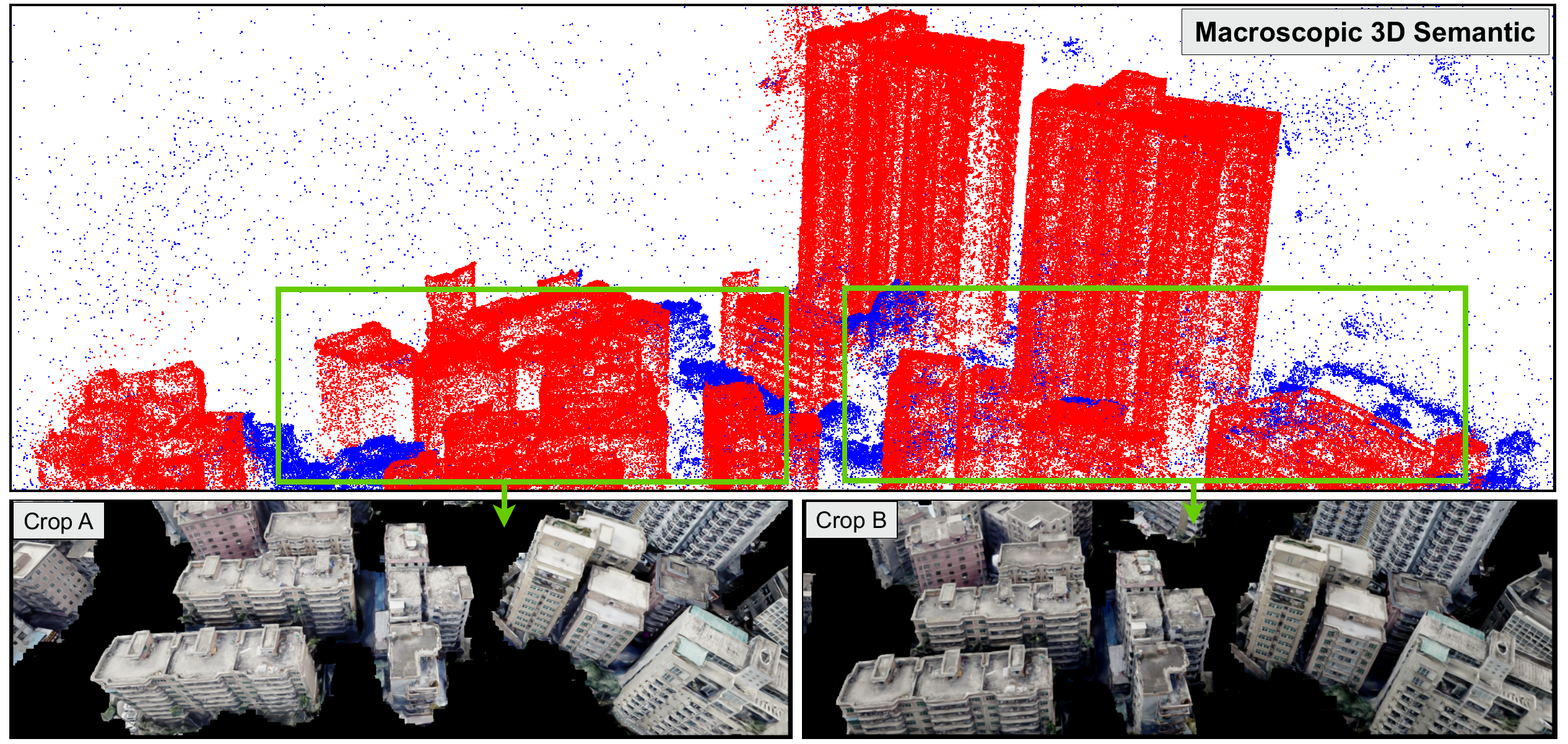}
    \caption{Large-scale automatic semantic segmentation results of PointGauss on the Campus360 dataset without human prompts. The macroscopic view demonstrates its capacity for urban mapping, while the zoom-in crops highlight the fine-grained preservation of architectural boundaries.}
    \label{fig:urbanSem}
\end{figure*}

\begin{figure*}
    \centering
    \includegraphics[width=\linewidth]{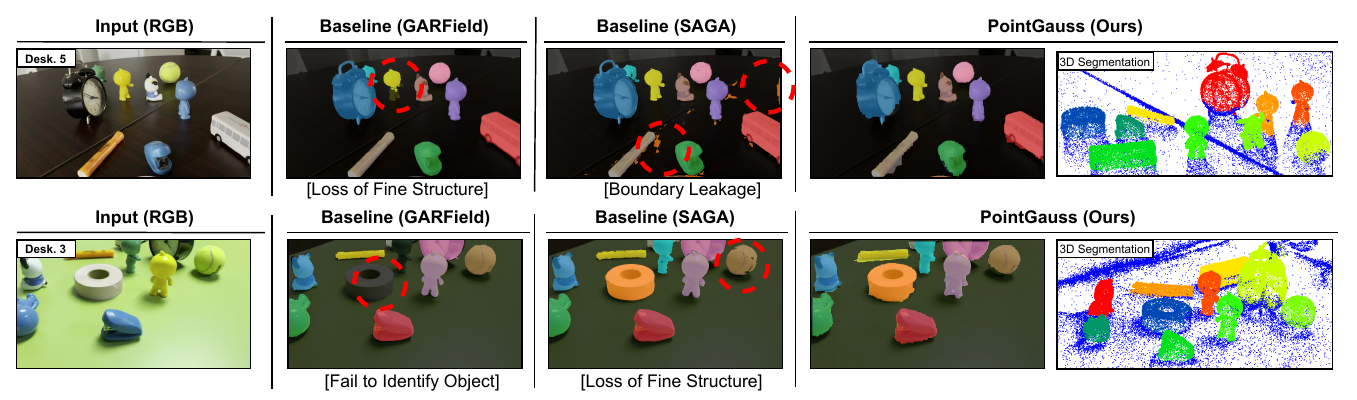}
    \caption{Qualitative comparison of instance segmentation on the \textit{Desk3} and \textit{Desk5} scenes within the \textit{DesktopObjects360} dataset. While 2D-based baselines like GARField and SAGA frequently suffer from multi-view inconsistency and partial segmentation errors (indicated by the \textbf{red dashed circles}), our PointGauss always maintains complete mask integrity and provides geometrically consistent 3D segmentation results.}
    \label{fig:deskInst}
\end{figure*}

\section{Experiments}
\label{sec:experiments}
In this section, we evaluate the performance of the proposed method on SplatSeg-360 benchmark, focusing on segmentation accuracy, algorithm deployment and execution efficiency, and overall robustness.

\subsection{Experimental Setup}
Data Splits. We evaluate our method and all baselines on the newly proposed SplatSeg-360 benchmark (detailed in Sec.~\ref{sec:dataset}). For a thorough evaluation, we split the scenes within each sub-dataset (Residence360, Campus360, Vehicles360, and DesktopObjects360) into training and testing sets, roughly following an 80\% / 20\% ratio. Due to the discrete number of scenes in certain sub-datasets, an exact 4:1 split is not always feasible. Specifically, our 3D segmentation network is trained on the training scenes and subsequently evaluated on the remaining unseen testing scenes.


Baselines. We compare PointGauss against four state-of-the-art 3DGS-based segmentation frameworks: Feature3DGS~\cite{feature3dgs}, OmniSeg-3DGS~\cite{omniseg}, SAGA~\cite{SAGAgaussian}, and GARField~\cite{garfield}. For a fair comparison, all baseline models are trained and evaluated on our benchmark using their official open-source implementations and recommended hyperparameters.

Evaluation Metrics. To comprehensively assess the segmentation quality, we employ three key metrics: 2D mean Intersection over Union (2D mIoU) for view-dependent mask accuracy, 3D Intersection over Union (3D IoU) for measuring semantic consistency directly on 3D Gaussian primitives, and Overall Accuracy (OA). The 3D IoU is particularly essential for verifying the model's true 3D spatial understanding.

Implementation Details. Our framework is implemented in PyTorch and built upon the original 2DGS ecosystem. PointGauss is optimized using the Adam optimizer with an initial learning rate of 1e-3. All experiments are conducted on a Linux server equipped with an 11th Gen Intel(R) Core(TM) i5-11400F CPU @ 2.60GHz, 64 GB of RAM, and a single NVIDIA GeForce RTX 4090 GPU.

\begin{table*}[t]
\caption{Quantitative Comparison of Instance Segmentation Performance on the Residence360 Sub-dataset. The Best Results are Highlighted in \textbf{Bold}. 3D represents 3D IoU, 2D represents 2D mIoU, and OA denotes Overall Accuracy.}
\label{tab:urban_results}
\centering
\small
\setlength{\tabcolsep}{2.8pt} 
\begin{tabular}{l ccc ccc ccc ccc ccc}
\toprule
\textbf{Method} & \multicolumn{3}{c}{\textbf{Scene 1}} & \multicolumn{3}{c}{\textbf{Scene 2}} & \multicolumn{3}{c}{\textbf{Scene 3}} & \multicolumn{3}{c}{\textbf{Scene 4}} & \multicolumn{3}{c}{\textbf{Scene 5}} \\
\cmidrule(lr){2-4} \cmidrule(lr){5-7} \cmidrule(lr){8-10} \cmidrule(lr){11-13} \cmidrule(lr){14-16}
 & 3D & 2D & OA & 3D & 2D & OA & 3D & 2D & OA & 3D & 2D & OA & 3D & 2D & OA \\
\midrule
Feature3DGS~\cite{feature3dgs} & -- & 35.06 & 59.86 & -- & 33.43 & 63.29 & -- & 34.76 & 58.67 & -- & 43.48 & 68.12 & -- & 43.52 & 67.67 \\
OmniSeg3D~\cite{omniseg}   & -- & 51.38 & 65.40 & -- & 37.01 & 58.80 & -- & 31.62 & 53.87 & -- & 43.40 & 67.94 & -- & 38.89 & 66.11 \\
GARField~\cite{garfield}    & -- & 10.25 & 33.28 & -- & 12.56 & 42.51 & -- & 22.41 & 45.13 & -- & 15.61 & 41.50 & -- & 11.57 & 39.47 \\
SAGA~\cite{SAGAgaussian}        & -- & 62.61 & 75.83 & -- & 68.22 & 82.69 & -- & 53.31 & 69.49 & -- & 62.05 & 79.64 & -- & 77.40 & 89.44 \\
\midrule
\textbf{PointGauss} & \textbf{90.68} & \textbf{81.55} & \textbf{89.77} & \textbf{92.42} & \textbf{84.94} & \textbf{93.70} & \textbf{89.55} & \textbf{72.10} & \textbf{82.73} & \textbf{83.39} & \textbf{81.77} & \textbf{90.36} & \textbf{89.80} & \textbf{85.82} & \textbf{93.19} \\
\bottomrule
\end{tabular}
\end{table*}

\subsection{Performance across Scales}
\label{subsec:performance_scales}

\subsubsection{Urban Building Segmentation}
\label{subsubsec:urban_build}
We first evaluate the instance segmentation performance on the \textit{Residence360} sub-dataset, which features large-scale city architectural structures characterized by intricate geometries and severe cross-view occlusions. Since the competitive baselines (Feature3DGS~\cite{feature3dgs}, OmniSeg3D~\cite{omniseg}, GARField~\cite{garfield}, and SAGA~\cite{SAGAgaussian}) are fundamentally designed for interactive, 2D-driven instance segmentation, they lack the capability to directly group or constrain 3D spatial primitives. Consequently, their 3D segmentation metrics (3D-IoU) cannot be natively computed, as denoted by ``--'' in Table~\ref{tab:urban_results}. 

As reported in Table~\ref{tab:urban_results}, the quantitative comparison demonstrates that the proposed PointGauss consistently and substantially outperforms all baseline methods across all five urban scenes in both 2D mask accuracy and overall scene understanding. SAGA emerges as the strongest baseline among the interactive frameworks, yet PointGauss yields remarkable performance gains over it. Specifically, PointGauss improves the 2D mIoU by $18.94\%$, $16.72\%$, $18.79\%$, $19.72\%$, and $8.42\%$ from Scene 1 to Scene 5, respectively. Similar scaling trends are observed in Overall Accuracy (OA), where PointGauss delivers a steady margin of improvement (up to $+13.94\%$ in Scene 1).

Crucially, baseline methods are architecturally limited to view-dependent 2D evaluations. In contrast, PointGauss achieves robust primitive-level 3D segmentation, maintaining a high 3D-IoU between 83.39\% and 92.42\%. This superior performance underscores the efficacy of our 3D-aware framework, which successfully injects geometric consistency into the Gaussian primitives. Consequently, it resolves the boundary ambiguities and multi-view inconsistencies that typically plague 2D interactive methods in complex urban scenarios. Fig.~\ref{fig:urbanInst} (a) visualizes the comparison between the best baseline (SAGA) and our method, where SAGA tends to yield incomplete segmentations for large-scale houses.

\begin{table}[t] 
\caption{Automatic Semantic Segmentation Performance of PointGauss on the Campus360 Subdataset. 3D represents 3D IoU, 2D represents 2D mIoU, and OA denotes Overall Accuracy. (Competitive Baselines are Omitted as They Do Not Support Automatic Inference).}
\label{tab:semantic_results}
\centering
\small
\setlength{\tabcolsep}{3pt} 
\begin{tabular}{l cc cc ccc} 
\toprule
\textbf{Scene} & \multicolumn{2}{c}{\textbf{Building}} & \multicolumn{2}{c}{\textbf{Background}} & \multicolumn{3}{c}{\textbf{ALL (Overall)}} \\
\cmidrule(lr){2-3} \cmidrule(lr){4-5} \cmidrule(lr){6-8}
 & 3D & 2D & 3D & 2D & 3D & 2D & OA \\
\midrule
Campus10 & 75.28 & 75.69 & 83.14 & 88.09 & 79.21 & 81.89 & 91.31 \\
Campus11 & 94.04 & 88.97 & 76.84 & 91.09 & 85.44 & 90.03 & 94.82 \\
\bottomrule
\end{tabular}
\end{table}

\subsubsection{Automatic Semantic Segmentation on Campus Scenes}
\label{subsubsec:semantic_seg}
Beyond interactive instance segmentation, a distinctive advantage of the proposed PointGauss is its capacity for \textit{automatic semantic segmentation} without requiring any manual prompts or human-in-the-loop interventions. Crucially, all competitive baseline methods (Feature3DGS, OmniSeg3D, GARField, and SAGA) are bound to interactive paradigms and are fundamentally incapable of conducting automatic, category-wide semantic inference. Thus, they fail to deliver valid results on this standard semantic benchmarking task. 

To verify this capability in practical large-scale environments, we evaluate PointGauss on two distinct campus areas (\textit{Campus Area 11} and \textit{Campus Area 12}), focusing on the automatic parsing of critical land-cover categories, namely Building and Background (encompassing ground and sky). As summarized in Table~\ref{tab:semantic_results}, PointGauss achieves exceptional performance in both 2D and 3D semantic spaces. For the target architectural structures, it yields a 3D-IoU of $75.28\%$ in Campus Area 11, which scales up to an impressive $94.04\%$ in Campus Area 12, demonstrating its high fidelity in delineating complex geometry boundaries. For the background, the 2D-IoU reaches up to $91.09\%$, reflecting strong generalization across complex textured terrains. Fig.~\ref{fig:urbanSem} illustrates the capability of segmenting buildings of our method. Overall, PointGauss achieves a high 3D-mIoU of $85.44\%$ and an Overall Accuracy (OA) of $94.82\%$ in Campus Area 12. These results firmly validate that our method not only excels at fine-grained object grouping but also serves as a robust tool for fully automated, large-scale urban semantic mapping.

\begin{table}[h]
\caption{Quantitative Evaluation of Vehicle Instance Segmentation on the Vehicles360 Subdataset. The Best Results are Highlighted in \textbf{Bold}. 3D represents 3D IoU, 2D represents 2D mIoU, and OA denotes Overall Accuracy.}
\label{tab:road_results}
\centering
\footnotesize
\setlength{\tabcolsep}{5pt}
\begin{tabular}{l|ccc|ccc}
\toprule
\textbf{Method} & \multicolumn{3}{c|}{\textbf{California3}} & \multicolumn{3}{c}{\textbf{TurkSt5}} \\
 & 3D & 2D & OA & 3D & 2D & OA \\
\midrule
Feature3DGS~\cite{feature3dgs} & -- & 70.46 & 94.97 & -- & 68.66 & 94.15 \\
OmniSeg3D~\cite{omniseg}   & -- & 60.37 & 90.47 & -- & 68.89 & 92.84 \\
GARField~\cite{garfield}    & -- & 80.40 & 96.27 & -- & 84.80 & 97.58 \\
SAGA~\cite{SAGAgaussian}        & -- & \textbf{96.23} & \textbf{99.41} & -- & \textbf{95.71} & \textbf{99.33} \\
\midrule
\textbf{PointGauss} & \textbf{91.50} & 93.14 & 98.71 & \textbf{91.38} & 91.77 & 98.42 \\
\bottomrule
\end{tabular}
\end{table}

\subsubsection{Medium-Scale Road Vehicle Segmentation}
\label{subsubsec:road_vehicle}
To evaluate the capability of our model in parsing medium-sized foreground targets within urban environments, we conduct experiments on the \textit{Vehicles360} sub-dataset, which explicitly captures road scenes with multiple vehicles. For this evaluation, eight street-view scenes are selected for training the network, while two representative scenes, namely \textit{California3} and \textit{TurkSt5}, are withheld for testing multi-view instance segmentation. 

The quantitative results are summarized in Table~\ref{tab:road_results}, and the qualitative visual comparisons are illustrated in Fig.~\ref{fig:urbanInst} (b). Among the existing continuous competitive frameworks, SAGA serves as the most competitive baseline in 2D evaluation, yielding the top-tier 2D mIoU and OA. As shown in the visual results of Fig.~\ref{fig:urbanInst} (b), although our proposed PointGauss always successfully segments the entire object mask as a whole, it exhibits slightly lower boundary precision compared to SAGA. This subtle limitation in edge delineation leads to a minor performance gap in 2D metrics, where PointGauss achieves a 2D mIoU of $93.14\%$ on \textit{California3} and $91.77\%$ on \textit{TurkSt5}, slightly trailing SAGA. 

However, 2D evaluation alone fails to reflect the underlying geometric robustness. In practice, 2D-based methods like SAGA suffer from multi-view inconsistency and partial viewpoint segmentation errors due to the lack of spatial constraints. In sharp contrast, because PointGauss directly executes segmentation within the 3D space, it theoretically and fundamentally eliminates the multi-view inconsistency problem entirely. 

More importantly, while existing methods yield blank outputs for 3D primitive evaluation due to their inherent procedural limitations, PointGauss delivers exceptional 3D-IoU scores exceeding $91.3\%$ ($91.50\%$ on \textit{California3} and $91.38\%$ on \textit{TurkSt5}). This demonstrates that PointGauss successfully overcomes the common cross-view identity-switching problem often encountered when segmenting movable specular objects like vehicles, providing highly consistent, stable, and geometrically sound 3D Gaussian primitive clusters.

\begin{table*}[t]
\centering
\caption{Experimental results on DesktopObjects360 sub-dataset (\%). 3D represents 3D IoU, 2D represents 2D mIoU, and OA denotes Overall Accuracy.}
\label{tab:seg_results_combined}
\small
\setlength{\tabcolsep}{2.0pt} 
\begin{tabular}{l ccc ccc ccc ccc ccc ccc}
\toprule
\multirow{2}{*}{Method} & 
\multicolumn{3}{c}{Desk.1} & \multicolumn{3}{c}{Desk.2} & \multicolumn{3}{c}{Desk.3} &
\multicolumn{3}{c}{Desk.4} & \multicolumn{3}{c}{Desk.5} & \multicolumn{3}{c}{Desk.6} \\
\cmidrule(lr){2-4} \cmidrule(lr){5-7} \cmidrule(lr){8-10} \cmidrule(lr){11-13} \cmidrule(lr){14-16} \cmidrule(lr){17-19}
 & 3D & 2D & OA & 3D & 2D & OA & 3D & 2D & OA & 3D & 2D & OA & 3D & 2D & OA & 3D & 2D & OA \\
\midrule
Feature3DGS~\cite{feature3dgs}  & -- & 38.36 & 71.49 & -- & 32.72 & 62.48 & -- & 42.27 & 71.45 & -- & 34.49 & 70.43 & -- & 32.54 & 66.16 & -- & 19.65 & 62.42 \\
OmniSeg3D~\cite{omniseg}        & -- & 41.19 & 49.60 & -- & 54.96 & 85.13 & -- & 47.40 & 78.27 & -- & 50.02 & 75.32 & -- & 58.20 & 89.49 & -- & 63.29 & 93.89 \\
GARField~\cite{garfield}        & -- & 66.22 & 92.52 & -- & 27.27 & 82.56 & -- & 71.48 & 91.86 & -- & 64.63 & 93.32 & -- & 58.82 & 90.61 & -- & 65.21 & 97.44 \\
SAGA~\cite{SAGAgaussian}        & -- & 68.72 & 85.67 & -- & 66.55 & 86.86 & -- & 69.42 & 87.31 & -- & 87.47 & 97.52 & -- & 79.40 & 95.41 & -- & 59.34 & 92.83 \\
\midrule
\textbf{PointGauss(Ours)}       & \textbf{69.40} & \textbf{84.33} & \textbf{95.90} & \textbf{82.46} & \textbf{86.91} & \textbf{96.67} & \textbf{73.38} & \textbf{85.85} & \textbf{94.54} & \textbf{78.05} & \textbf{89.36} & \textbf{97.68} & \textbf{80.94} & \textbf{85.55} & \textbf{96.76} & \textbf{77.67} & \textbf{91.12} & \textbf{99.31} \\
\bottomrule
\end{tabular}
\end{table*}

\subsubsection{Small-Scale Tabletop Object Segmentation}
\label{subsubsec:desktop_objects}
Finally, to assess the adaptability and localized precision of our framework on micro-scale targets, we evaluate PointGauss on the \textit{DesktopObjects360} benchmark. This sub-dataset consists of six close-up tabletop environments (\textit{Desk.1} to \textit{Desk.6}) that pose severe challenges to 3DGS segmentation, including small spatial occupancy, fine-grained object boundaries, structural proximity, and complex object-to-object occlusions. 

As reported in Table~\ref{tab:seg_results_combined}, PointGauss consistently achieves the highest performance across all six scenarios, establishing its overwhelming superiority in fine-grained scene parsing. This quantitative superiority is further validated by the qualitative results illustrated in Fig.~\ref{fig:deskInst}, which contrasts our method against SAGA and GARField—the two top-performing baselines on this benchmark—across the \textit{Desk.3} and \textit{Desk.5} scenes. As manifested in the visual comparisons, these state-of-the-art 2D-driven baselines still frequently suffer from incomplete object masks or catastrophic segmentation failures when dealing with intricate structures. In sharp contrast, our proposed PointGauss consistently achieves meticulous and precise object-level segmentation. 

Such robust visual fidelity perfectly aligns with the statistical metrics. While competitive interactive baselines like SAGA and GARField deliver acceptable results in specific scenes (e.g., SAGA yields an $87.47\%$ 2D mIoU on \textit{Desk.4}), their performance degrades sharply in more complex layouts due to the lack of explicit 3D spatial constraints. In contrast, PointGauss exhibits exceptional robustness; most notably, on \textit{Desk.6}, our method elevates the 2D mIoU to an impressive $91.12\%$, outperforming SAGA ($59.34\%$) and GARField ($65.21\%$) by a substantial margin of $+31.78\%$ and $+25.91\%$, respectively. Significant gains are also observed in Overall Accuracy (OA), where PointGauss achieves a near-perfect parsing accuracy of $99.31\%$ on \textit{Desk.6}.

Crucially, PointGauss successfully delivers high-fidelity 3D-IoU scores across all scenes, peaking at $82.46\%$ on \textit{Desk.2}. Unlike the alternative methods that suffer from severe boundary bleeding and identity disintegration when handling small items, our 3D-aware framework maintains sharp geometric delineations directly on the Gaussian primitives. This compelling performance across macro-scale (urban buildings), meso-scale (road vehicles), and micro-scale (tabletop items) environments convincingly demonstrates that PointGauss possesses exceptional multi-scale versatility and robust geometric-semantic consistency.

\begin{table}[t]
\caption{3D preparation (3D Prep.) and per-frame inference time comparison. * indicates the time of post-processing}
\label{tab:time_efficiency}
\centering
\begin{tabular}{lcc}
\toprule
\textbf{Method} & \textbf{3D Prep. (min)} & \textbf{Per-Frame (ms)} \\
\midrule
GARField~\cite{garfield}        & 45 & 3232 \\
OmniSeg3D~\cite{omniseg}         & 37 & 463 \\
Feature3DGS~\cite{feature3dgs}     & 35 & 510 \\
SAGA~\cite{SAGAgaussian}            & 32 & \textbf{31} \\
PointGauss (Ours)   & \textbf{0.13} & 5+388* \\
\bottomrule
\end{tabular}

\end{table}


\begin{figure}[t]
    \centering
    \includegraphics[width=\linewidth]{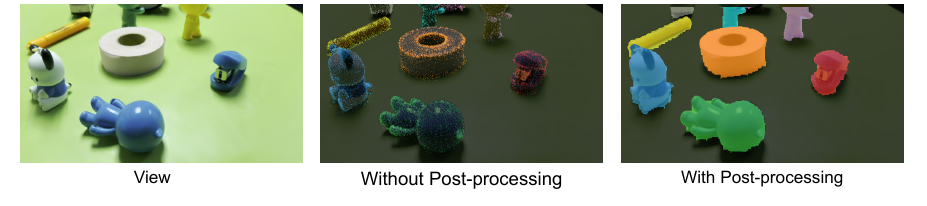}
    \caption{Post-processing Module Analysis. The middle part, without post-processing, exhibits instance fragmentation (scattered colorful points). The right part, with our post-processing, maintains instance integrity (solid-colored regions). }
    \label{fig:postprocessing}
\end{figure}
\subsection{Time Efficiency Analysis}
We evaluated the time efficiency of SAGA, Feature3DGS, OmniSeg3D, GARField, and PointGauss on the Desk. 6 scene of the DesktopObjects360 sub-dataset, focusing on 3D information preparation and per-frame processing. Unlike existing baselines that rely on pre-trained 2D foundation models (e.g., SAM), PointGauss utilizes point cloud segmentation models. Since the trained model can be reused across multiple Gaussian scenes, we exclude its training time (approximately 9 minutes) when comparing computational efficiency on individual scenes.

As shown in Table~\ref{tab:time_efficiency}, PointGauss achieves near-instant 3D scene preparation in 0.13 minutes, compared to 32-45 minutes for other methods—a 200× to 300× speedup. This eliminates the need for offline preprocessing, enabling real-time interaction. For per-frame inference, SAGA excels with 31 ms, while PointGauss requires 5 ms (core inference) + 388 ms (post-processing). Although post-processing introduces some latency, our core inference is highly competitive (e.g., 6× faster than SAGA).

Our efficiency stems from directly operating on Gaussian primitives, while other methods must process hundreds of input images during the preparation stage.

\begin{table}[t]
\caption{Effect of the Number of Clicks on Segmentation Performance}
\label{tab:click_ablation}
\centering
\begin{tabular}{cccc}
\toprule
\textbf{Num. of Clicks} & \textbf{3D IoU (\%)} & \textbf{2D mIoU (\%)} & \textbf{OA (\%)} \\
\midrule
5  & 68.72 & 90.01 & 99.22 \\
10 & 73.29 & 90.99 & 99.29 \\
15 & 75.17 & 90.80 & 99.29 \\
20 & 77.67 & 91.12 & 99.31 \\
25 & 80.61 & 91.18 & 99.30 \\
30 & 81.83 & 90.95 & 99.29 \\
\bottomrule
\end{tabular}
\vspace{-0.2cm}
\end{table}

\subsection{Ablation Experiments}
\subsubsection{Impacts of Clicks}  
To assess the impact of click quantity on segmentation performance, we conduct experiments on Desk. 6 scene of the DesktopObjects360 su-bdataset with varying number of clicks. Table~\ref{tab:click_ablation} shows that increasing the number of clicks improves 3D IoU (from 68.72\% to 81.83\%), with diminishing returns beyond 25 clicks. In contrast, 2D mIoU remains relatively stable (~91\%), indicating that 2D segmentation requires fewer clicks. Overall Accuracy (OA) stays consistently high ($>$ 99.2\%), demonstrating robustness to variations in click count. Optimal performance is observed with 20-25 clicks.

\subsubsection{Impacts of Post-processing}
To evaluate the impact of post-processing, we conduct an ablation study (Fig.~\ref{fig:postprocessing}). Without post-processing, the segmentation mask contains holes and artifacts, leading to incomplete and noisy rendered outputs. In contrast, applying post-processing results in smoother and more complete masks, producing clearer and more accurate segmentation. These results highlight the essential role of post-processing in enhancing segmentation quality.


\section{Conclusion}
\label{sec:conclusion}
In this article, we presented \textit{PointGauss}, a unified, geometry-centric 3D-native framework that addresses the critical geometric inconsistencies and scalability bottlenecks of traditional 2D-to-3D distillation paradigms for 3D Gaussian Splatting (3DGS) scene understanding. By leveraging Point Transformer V3 (PTv3) to extract scale-invariant features directly from unstructured Gaussian attributes, PointGauss inherently eliminates projection ambiguities across complex urban topographies. Furthermore, our adaptive region-of-interest (ROI) cropping strategy and instance-aware, distance-constrained rasterization pipeline significantly mitigate computational overheads, ensuring rapid semantic initialization and pixel-level, view-consistent label propagation. To bridge the evaluation gap, we also introduced \textbf{SplatSeg-360}, the first comprehensive, cross-scale benchmark explicitly tailored for 3DGS, providing over 6,300 natively aligned 2D-3D masks across 32 diverse 360$^\circ$ scenes. Extensive evaluations demonstrate that PointGauss operates in real-time and achieves state-of-the-art performance, yielding an exceptional 90\% 3D-mIoU in building-scale scenarios and roughly 80\% 2D-mIoU in 2D instance segmentation. Future work will extend this framework to dynamic environments and explore multi-modal remote sensing data fusion to further advance intelligent digital twin modeling.

\bibliographystyle{IEEEtran}
\bibliography{main}

 

\newpage

 





\end{document}